%% file: main.tex
\documentclass[]{style/ceurart}
\usepackage{listings}
\begin{document}

\copyrightyear{2025}
\copyrightclause{Copyright for this paper by its authors.
  Use permitted under Creative Commons License Attribution 4.0
  International (CC BY 4.0).}

\conference{CLEF 2025 Working Notes, 9 -- 12 September 2025, Madrid, Spain}

\title{LLM-Guided Planning and Summary-Based Scientific Text Simplification: DS@GT at CLEF 2025 SimpleText}

\author[1]{Krishna Chaitanya Marturi}[
    email=kmarturi3@gatech.edu,
]
\cormark[1]
\author[2]{Heba H. Elwazzan}[
    email=helwazzan3@gatech.edu
]

\address[1]{Georgia Institute of Technology, North Ave NW, Atlanta, GA 30332}

\cortext[1]{Corresponding author.}

\begin{abstract}
    In this paper, we present our approach for the CLEF 2025 SimpleText Task 1, which addresses both sentence-level and document-level scientific text simplification. For sentence-level simplification, our methodology employs large language models (LLMs) to first generate a structured plan, followed by plan-driven simplification of individual sentences. At the document level, we leverage LLMs to produce concise summaries and subsequently guide the simplification process using these summaries. This two-stage, LLM-based framework enables more coherent and contextually faithful simplifications of scientific text.
\end{abstract}


\maketitle

% We recommend splitting your main document into smaller parts that are easier to navigate.
% The input command "imports" the contents of the file into the current location.
% Prefixing the document with a number allows for natural string sorts.
\input{sections/00_main}

\bibliography{main}

\appendix

\section{Codabench Competition Submissions}

\subsection{Submissions for Task 1.1}
\label{app:codabench-task1.1}
\begin{table}[h!]
\centering
\caption{Codabench Submission Details for SimpleText Task 1.1 (Sentence-Level Simplification)}
\label{tab:codabench-task11-extra}
\begin{tabular}{|l|c|l|l|}
\hline
\textbf{Competition} & \textbf{ID \#} & \textbf{File Name} & \textbf{Task} \\
\hline
SimpleText Task 1 & 321007 & \texttt{dsgt\_Task11\_llama\_simplifier.zip} & Task 1.1 Sentence Level \\
SimpleText Task 1 & 303306 & \texttt{dsgt\_Task11\_plan\_guided\_llama.zip} & Task 1.1 Sentence Level \\
\hline
\end{tabular}
\end{table}

\subsection{Submissions for Task 1.2}
\label{app:codabench-task1.2}

\begin{table}[h!]
\centering
\caption{Codabench Submission Details for SimpleText Task 1.2 (Document-Level Simplification)}
\label{tab:codabench-task12-extra}
\begin{tabular}{|l|c|l|l|}
\hline
\textbf{Competition} & \textbf{ID \#} & \textbf{File Name} & \textbf{Task} \\
\hline
SimpleText Task 1 & 321021 & \texttt{dsgt\_Task12\_llama\_simplification.zip} & Task 1.2 Document Level \\
SimpleText Task 1 & 303443 & \texttt{dsgt\_Task12\_llama\_summary\_simplific.zip} & Task 1.2 Document Level \\
\hline
\end{tabular}
\end{table}

\section{Prompt Templates}

\subsection{LLM Prompt for Plan-Driven Sentence Simplification}
\label{app:prompt-plan-driven-simplification}

\begin{lstlisting}
You are a sentence simplifier.
Given a document, a sentence from that document, and the next sentence for context, choose an internal simplification strategy from the following options:
'rephrase', 'delete', 'split', 'ignore', 'merge'.
Then output ONLY the simplified sentence, based on your chosen strategy.

Document: The economic report showed a significant downturn in the last quarter.
Sentence: The economic report showed a significant downturn in the last quarter.
Next Sentence: Unemployment rates also rose sharply.
Simplified: The report said the economy got worse last quarter.

Document: Online social media provide users with unprecedented opportunities to engage with diverse opinions.
Sentence: Online social media provide users with unprecedented opportunities to engage with diverse opinions.
Next Sentence: They also enable misinformation to spread quickly.
Simplified: Social media let people easily share their opinions.

Document: We included seven cluster-randomised trials with 42,489 patient participants from 129 hospitals, conducted in Australia, the UK, China, and the Netherlands. Health professional participants (numbers not specified) included nursing, medical and allied health professionals. Interventions in all studies included implementation strategies targeting healthcare workers; three studies included delivery arrangements, no studies used financial arrangements or governance arrangements. Five trials compared a multifaceted implementation intervention to no intervention, two trials compared one multifaceted implementation intervention to another multifaceted implementation intervention. No included studies compared a single implementation intervention to no intervention or to a multifaceted implementation intervention. Quality of care outcomes (proportions of patients receiving evidence-based care) were included in all included studies. All studies had low risks of selection bias and reporting bias, but high risk of performance bias. Three studies had high risks of bias from non-blinding of outcome assessors or due to analyses used.
Sentence: We included seven cluster-randomised trials with 42,489 patient participants from 129 hospitals, conducted in Australia, the UK, China, and the Netherlands.
Next Sentence: Health professional participants (numbers not specified) included nursing, medical and allied health professionals.
Simplified:
\end{lstlisting}

\subsection{LLM Prompt for Document Summarization}
\label{app:prompt-document-summarization}

\begin{lstlisting}
You are given a complex document. Your task is to write a clear and concise summary that captures the essential information, main arguments, and key findings.

Guidelines:
- Do not include minor details or examples unless crucial to the main idea.
- Focus on the overall message and structure of the document.
- Use simple and accessible language.
- The summary should be understandable without reading the original document.

### Document:
{document}

### Summary:
\end{lstlisting}

\subsection{LLM Prompt for Summary-Guided Document Simplification}
\label{app:prompt-summary-guided-simplification}

\begin{lstlisting}[language=, caption={Prompt for LLM-Based Summary-Guided Document Simplification}]
You are given a complex document and its summary. Your task is to rewrite the complex document in a simpler, clearer way while ensuring the meaning aligns with the provided summary.

Guidelines:
- Keep the rewritten version faithful to both the original document and its summary.
- Use simple, accessible vocabulary and sentence structures.
- Avoid introducing new information not present in the original document.
- Retain the key ideas, structure, and intent captured in the summary.

### Complex Document:
{document}

### Summary:
{summary}

### Simplified Document:
\end{lstlisting}

\end{document}

%% file: sections/00_main.tex
\section{Introduction} \label{introduction}

In the past decade or so, a new form of learning has emerged. Instead of science being only accessible through journals or formal education, the general public is now privy to an enormous wealth of material through the Internet. Whether it be directed self-studying, or casual social media perusal, nearly everyone is now able to access scientific information. An important caveat remains, however, and that is when a resource is accessible for free and without rigorous moderation, misinformation will invariably run rampant. This is why now more than ever, the need for reliable, easy-to-understand scientific-based text and content has taken the forefront. 

This is where automated text simplification comes in. With the volume of scientific text at hand, rewriting the same exact content in layman terms manually is intractable. Resources have been expended towards automating this task, and over the course of 20 years, automatic text simplification has progressed significantly \cite{saggion2022automatic}, reaching a critical point with the development of Natural Language processing techniques, and more recently, with the widespread use of LLMs.

The capabilities of Large Language Models have made them a game-changer for automatic text simplification. Unlike previous methods, LLMs can achieve a deeper semantic interpretation of source text, allowing them to not only simplify vocabulary and syntax but also to summarize and restructure information for clarity. Though their internal representations of knowledge are opaque, their practical application as a powerful tool for generating simplified text is undeniable, paving the way for more effective simplification systems.

The SimpleText lab \cite{Ermakova2025Overview} is part of CLEF, and it aims to address the task of text simplification of scientific text. Task 1 in particular \cite{Bakker2025Overview} involves investigating the performance of text simplification on both the sentence-level and document-level. It uses the Cochrane-Auto dataset \cite{bakker-kamps-2024-cochrane} and uses standard text simplification metrics such as SARI, BLEU, BERTscore, etc. to evaluate the generated simplified text against the reference ones.

In this paper, we tackle both sentence-level text simplification as well as document-level using a tiered approach. For the sentence-level, an LLM first generates a simplification strategy regarding a sentence, and then the LLM is tasked to perform that strategy to generate a simplified version. For the document-level, the LLM generates a summary of the text as a whole and then uses that summary as part of the prompt that guides the simplification process.

The paper is organized as follows: Section \ref{related work} provides a small literature review of the text simplification task; section \ref{methodology} provides details of the approach undertaken for each of the subtasks; section \ref{results} showcases the results and provides a brief discussion; section \ref{conclusions} discusses possible future work and the conclusions we have derived from these experiments.

\section{Related Work} \label{related work}

Scientific text simplification aims to enhance the accessibility and comprehensibility of technical content for non-expert audiences, including patients, educators, and the general public. This task has been explored at both sentence and document levels, with increasing interest in using neural and large language model (LLM)-based methods.

\citet{ondov2022survey} provide a comprehensive survey of automated methods for biomedical text simplification. They categorize approaches into rule-based, statistical, and neural systems, highlighting the trade-offs between linguistic control and generative fluency. Their analysis underscores challenges in maintaining factual consistency and domain-specific accuracy, particularly in biomedical domains. This motivates the need for more grounded, interpretable approaches, such as plan-driven or summary-guided simplification, which we explore in this paper.

Recent work has also examined the role of LLMs in improving user comprehension and reducing cognitive load. \citet{guidroz2025llmbasedtextsimplificationeffect} evaluate how LLM-based simplification affects the understanding of the readers and the mental effort of different audiences. They find that while LLM-generated simplifications generally improve readability, there is a risk of hallucination and over-simplification, especially in scientific and biomedical texts.

\citet{fang2025progds} propose a hierarchical strategy involving LLMs to simplify document level sentences using a progressive process, that breaks it down to discourse-level,  topic-level, and lexical-level simplification. The task is formulated as a conditional generation problem by autoregressively conditioning on the input source document. This approach effectively preserves the content of the document while eliminating ambiguity and subjectivity, and avoids treating the document simplification task as merely document summarization. 

These findings support our motivation to investigate structured prompting methods to enhance control, coherence, and factuality in the simplification of scientific texts.

\subsection{Evaluation Metrics}

A variety of automatic evaluation metrics are employed to assess the quality of generated or simplified text. In this work, we consider four commonly used metrics: SARI, BLEU, BERTScore (F1) and Flesch-Kincaid grade level (FKGL), each capturing different aspects of text quality to compare LLM-based simplification strategies.

\paragraph{SARI} (System output Against References and against the Input sentence)~\cite{xu-etal-2016-optimizing} is specifically designed for the text simplification task. Unlike traditional metrics, SARI compares the system output not only to reference simplifications but also to the original input. It evaluates the quality of three operations: \textit{keeping} relevant words, \textit{deleting} unnecessary ones, and \textit{adding} appropriate new content. SARI is computed as the average of F1 scores for these three operations and is typically scaled from 0 to 100, with higher scores indicating better simplification quality.

\paragraph{BLEU} (Bilingual Evaluation Understudy)~\cite{papineni2002bleu} is an n-gram precision-based metric widely used in machine translation and text generation. It measures the overlap between system outputs and reference texts, incorporating a brevity penalty to discourage overly short outputs. However, BLEU is less suitable for simplification, as it tends to penalize edits that diverge lexically from the reference even when such changes improve simplicity or meaning.

\paragraph{BERTScore}~\cite{zhang2019bertscore} evaluates the semantic similarity between the generated text and the reference using contextual embeddings from a pretrained transformer model. The F1 variant computes the harmonic mean of precision and recall based on cosine similarity between token embeddings. BERTScore is particularly useful in capturing semantic adequacy, especially when lexical overlap is low but the meaning is preserved.

\paragraph{Flesch–Kincaid Grade Level (FKGL)}~\cite{kincaid1975derivation} is a readability metric that estimates the U.S. school grade level required to comprehend the text. It is computed using average sentence length and average syllables per word. Lower FKGL scores indicate simpler text and are often used as a proxy for evaluating readability in simplification tasks. However, FKGL focuses solely on surface-level features and does not consider syntactic or semantic correctness \cite{tanprasert-kauchak-2021-flesch}.

\section{Methodology} \label{methodology}

The focus of task 1 is to study the performance of simplification systems in both sentence-level and document-level settings. 

\subsection{Sentence-level simplification - Task 1.1}
\begin{figure}[h]
    \centering
    \includegraphics[width=0.8\textwidth]{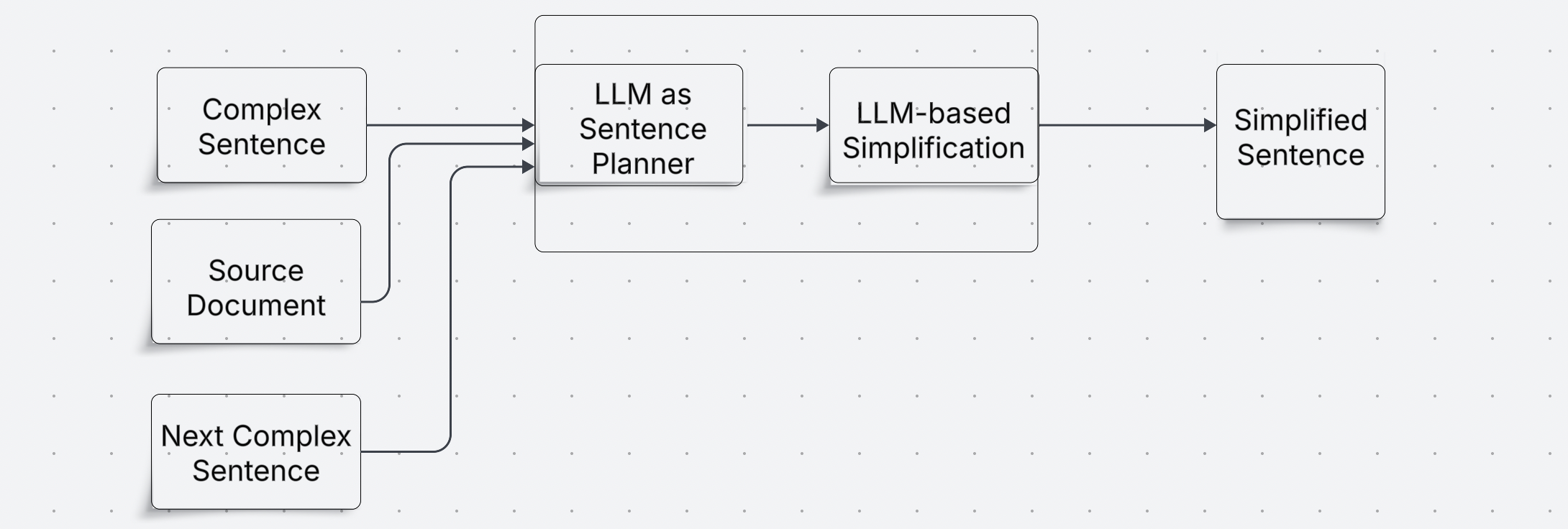} % adjust width as needed
    \caption{LLM based Plan-Driven Sentence Simplification - Task 1.1}
    \label{fig:task_1_1}
    \end{figure}

Inspired by recent advances in plan-driven sentence simplification~\cite{bakker-kamps-2024-cochrane}, we adopt a large language model (\texttt{llama-3.3-70b-versatile}) as a plan-based simplifier. As illustrated in Figure~\ref{fig:task_1_1}, this approach utilizes few-shot prompting with three inputs: a complex sentence, its corresponding source document, and the next complex sentence from the document.

The task is structured into two stages. In the first stage, the model is prompted to select an appropriate simplification strategy from a predefined set: \textit{rephrase}, \textit{delete}, \textit{split}, \textit{ignore}, or \textit{merge}. In the second stage, the model is prompted to generate the corresponding simplified sentence based on the selected strategy[Appendix~\ref{app:prompt-plan-driven-simplification}].

\subsection{Document-level simplification - Task 1.2}
\begin{figure}[h]
    \centering
    \includegraphics[width=0.8\textwidth]{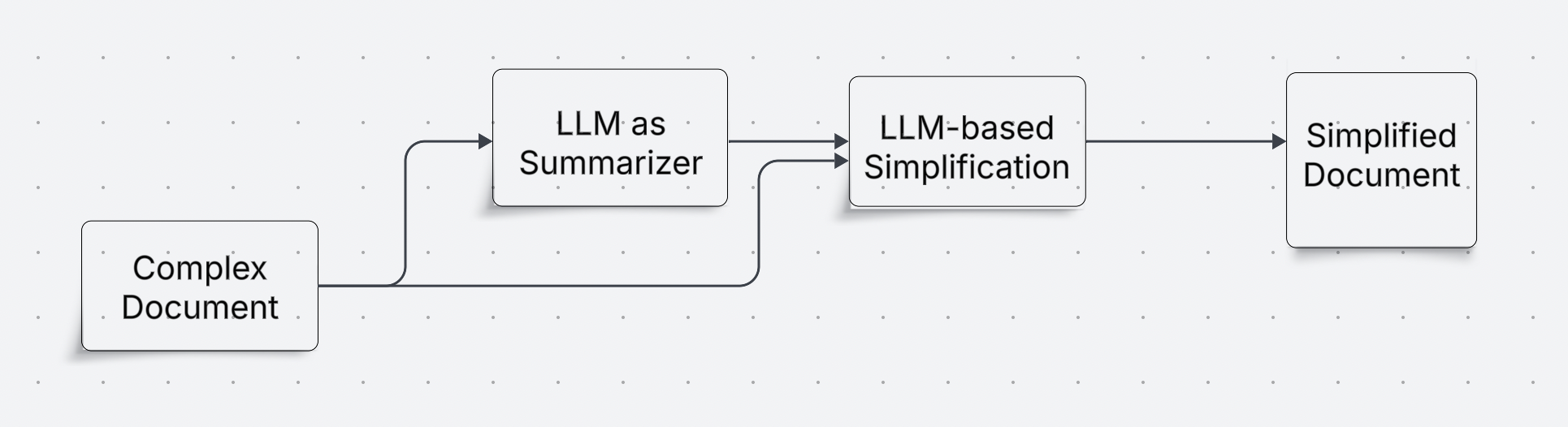} % adjust width as needed
    \caption{LLM based Summary-Guided Document Simplification - Task 1.2}
    \label{fig:task_1_2}
    \end{figure}

In this task, we leverage large language models (LLMs) for summary-guided document simplification~\cite{fang2025progds}.  Figure~\ref{fig:task_1_2} outlines our two-step pipeline, where a large language model, \texttt{llama-3.3-70b-versatile}, is employed both as a summarizer and a simplifier. 

First, the model is prompted to produce a clear and concise summary of the input complex document[Appendix~\ref{app:prompt-document-summarization}]. This summary serves as a semantic scaffold to guide the simplification process. In the second step, the same model is prompted to simplify the original document using the generated summary as contextual guidance[Appendix~\ref{app:prompt-summary-guided-simplification}]. This strategy enhances coherence and faithfulness while reducing the risk of over-simplification.

\section{Results} \label{results}

\subsection{Evaluation of Task 1.1: Sentence-level Scientific Text Simplification }

The \texttt{plan\_guided\_llama} system demonstrates effective sentence-level simplification on the 37 aligned Cochrane-auto abstracts, achieving a strong SARI score of 42.33 and reducing lexical complexity (8.52) while maintaining reasonable BLEU and FKGL values. This indicates that the model balances simplification and content preservation. Detailed results are shown in Table~\ref{tab:task1.1-dsgt}.

\begin{table}[h!]
\centering
\caption{Results for CLEF 2025 SimpleText Task 1.1 Sentence-Level Text Simplification: Test data on 37 aligned Cochrane-auto abstracts}
\label{tab:task1.1-dsgt}
\resizebox{\textwidth}{!}{%
\begin{tabular}{|l|c|c|c|c|c|c|c|c|c|c|c|}
\hline
\textbf{Method} & \rotatebox[origin=c]{90}{\textbf{Count}} & \rotatebox[origin=c]{90}{\textbf{SARI}} & \rotatebox[origin=c]{90}{\textbf{BLEU}} & \rotatebox[origin=c]{90}{\textbf{FKGL}} & \rotatebox[origin=c]{90}{\textbf{Compression Ratio}} & \rotatebox[origin=c]{90}{\textbf{Sentence Splits}} & \rotatebox[origin=c]{90}{\textbf{Levenshtein Similarity}} & \rotatebox[origin=c]{90}{\textbf{Exact Copies}} & \rotatebox[origin=c]{90}{\textbf{Additions Proportion}} & \rotatebox[origin=c]{90}{\textbf{Deletions Proportion}} & \rotatebox[origin=c]{90}{\textbf{Lexical Complexity Score}} \\
\hline
Source & 37 & 12.03 & 20.53 & 13.54 & 1.00 & 1.00 & 1.00 & 1.00 & 0.00 & 0.00 & 8.89 \\
Reference & 37 & 100.00 & 100.00 & 11.73 & 0.56 & 0.67 & 0.50 & 0.00 & 0.16 & 0.60 & 8.71 \\
plan\_guided\_llama & 37 & 42.33 & 10.43 & 7.77 & 0.48 & 0.97 & 0.47 & 0.00 & 0.18 & 0.70 & 8.52 \\
\hline
\end{tabular}
}
\end{table}

The \texttt{plan\_guided\_llama} model demonstrates strong simplification capability on the 217 Plain Language Summaries test set, achieving a SARI score of 42.98, which reflects balanced simplification performance. However, its BLEU score (6.33) is notably low, suggesting limited surface-level overlap with references. The model produces simplified outputs with a significantly lower FKGL (7.82) compared to the source (13.29), indicating improved readability. The compression ratio (0.48) and low exact copy rate (0.00) suggest aggressive simplification, while the deletions proportion (0.71) is higher than additions (0.18), pointing to a tendency to simplify by removal. Full results are shown in Table~\ref{tab:task1.1-dsgt-217}.

\begin{table}[h!]
\centering
\caption{Results for CLEF 2025 SimpleText Task 1.1 sentence-level text simplification: Test data on 217 Plain Language Summaries}
\label{tab:task1.1-dsgt-217}
\resizebox{\textwidth}{!}{
\begin{tabular}{|l|c|c|c|c|c|c|c|c|c|c|c|}
\hline
\textbf{Method} & \rotatebox[origin=c]{90}{\textbf{Count}} & \rotatebox[origin=c]{90}{\textbf{SARI}} & \rotatebox[origin=c]{90}{\textbf{BLEU}} & \rotatebox[origin=c]{90}{\textbf{FKGL}} & \rotatebox[origin=c]{90}{\textbf{Compression Ratio}} & \rotatebox[origin=c]{90}{\textbf{Sentence Splits}} & \rotatebox[origin=c]{90}{\textbf{Levenshtein Similarity}} & \rotatebox[origin=c]{90}{\textbf{Exact Copies}} & \rotatebox[origin=c]{90}{\textbf{Additions Proportion}} & \rotatebox[origin=c]{90}{\textbf{Deletions Proportion}} & \rotatebox[origin=c]{90}{\textbf{Lexical Complexity Score}} \\
\hline
Source & 217 & 7.84 & 10.55 & 13.29 & 1.00 & 1.00 & 1.00 & 1.00 & 0.00 & 0.00 & 9.05 \\
Reference & 217 & 100.00 & 100.00 & 11.28 & 0.72 & 0.97 & 0.40 & 0.00 & 0.29 & 0.63 & 8.65 \\
\texttt{plan\_guided\_llama} & 217 & 42.98 & 6.33 & 7.82 & 0.48 & 0.99 & 0.46 & 0.00 & 0.18 & 0.71 & 8.50 \\
\hline
\end{tabular}
}
\end{table}

Overall, the results demonstrate that the plan-guided LLaMA system effectively simplifies complex biomedical text at sentence-level while maintaining readability and informativeness, with a trade-off in exact lexical overlap.

\subsection{Evaluation of Task 1.2: Document-level Scientific Text Simplification}

The results in Table~\ref{tab:task1.2-dsgt-cochrane} show that the \texttt{llama\_summary\_simplification} system achieves a SARI score of 40.32, indicating moderate simplification quality. While the BLEU score (7.63) is comparatively low—suggesting some divergence from reference phrasing—the FKGL score of 9.56 reflects improved readability relative to the source. The compression ratio of 0.59 and deletion proportion of 0.70 suggest aggressive content reduction, contributing to simplification but potentially at the cost of semantic fidelity.

\begin{table}[h!]
\centering
\caption{Results for CLEF 2025 SimpleText Task 1.2 document-level text simplification on 37 aligned Cochrane-auto abstracts}
\label{tab:task1.2-dsgt-cochrane}
\resizebox{\textwidth}{!}{
\begin{tabular}{|l|c|c|c|c|c|c|c|c|c|c|c|}
\hline
\textbf{Method} & \rotatebox[origin=c]{90}{\textbf{Count}} & \rotatebox[origin=c]{90}{\textbf{SARI}} & \rotatebox[origin=c]{90}{\textbf{BLEU}} & \rotatebox[origin=c]{90}{\textbf{FKGL}} & \rotatebox[origin=c]{90}{\textbf{Compression Ratio}} & \rotatebox[origin=c]{90}{\textbf{Sentence Splits}} & \rotatebox[origin=c]{90}{\textbf{Levenshtein Similarity}} & \rotatebox[origin=c]{90}{\textbf{Exact Copies}} & \rotatebox[origin=c]{90}{\textbf{Additions Proportion}} & \rotatebox[origin=c]{90}{\textbf{Deletions Proportion}} & \rotatebox[origin=c]{90}{\textbf{Lexical Complexity Score}} \\
\hline
Source & 37 & 12.03 & 20.53 & 13.54 & 1.00 & 1.00 & 1.00 & 1.00 & 0.00 & 0.00 & 8.89 \\
Reference & 37 & 100.00 & 100.00 & 11.73 & 0.56 & 0.67 & 0.50 & 0.00 & 0.16 & 0.60 & 8.71 \\
llama\_summary\_simplification & 37 & 40.32 & 7.63 & 9.56 & 0.59 & 0.86 & 0.42 & 0.00 & 0.31 & 0.70 & 8.49 \\
\hline
\end{tabular}
}
\end{table}

The \texttt{llama\_summary\_simplification} method demonstrates effective simplification on the 217 Plain Language Summaries, achieving a strong SARI score of 42.92 and a reduced FKGL of 9.94. The lexical complexity score of 8.55 indicates simplification in vocabulary. However, a relatively low BLEU score of 5.32 and Levenshtein similarity of 0.39 suggest reduced semantic similarity to the reference (Table~\ref{tab:task1.2-dsgt-pls}).

Overall, the summary guided method achieves consistent simplification across both the datasets, balancing lower complexity with reduced semantic fidelity.

\begin{table}[h!]
\centering
\caption{Results for CLEF 2025 SimpleText Task 1.2 Document-Level Text Simplification: Test Data on 217 Plain Language Summaries }
\label{tab:task1.2-dsgt-pls}
\resizebox{\textwidth}{!}{
\begin{tabular}{|l|c|c|c|c|c|c|c|c|c|c|c|}
\hline
\textbf{Method} & \rotatebox[origin=c]{90}{\textbf{Count}} & \rotatebox[origin=c]{90}{\textbf{SARI}} & \rotatebox[origin=c]{90}{\textbf{BLEU}} & \rotatebox[origin=c]{90}{\textbf{FKGL}} & \rotatebox[origin=c]{90}{\textbf{Compression Ratio}} & \rotatebox[origin=c]{90}{\textbf{Sentence Splits}} & \rotatebox[origin=c]{90}{\textbf{Levenshtein Similarity}} & \rotatebox[origin=c]{90}{\textbf{Exact Copies}} & \rotatebox[origin=c]{90}{\textbf{Additions Proportion}} & \rotatebox[origin=c]{90}{\textbf{Deletions Proportion}} & \rotatebox[origin=c]{90}{\textbf{Lexical Complexity Score}} \\
\hline
Source & 217 & 7.84 & 10.55 & 13.29 & 1.00 & 1.00 & 1.00 & 1.00 & 0.00 & 0.00 & 9.05 \\
Reference & 217 & 100.00 & 100.00 & 11.28 & 0.72 & 0.97 & 0.40 & 0.00 & 0.29 & 0.63 & 8.65 \\
llama\_summary\_simplification & 217 & 42.92 & 5.32 & 9.94 & 0.49 & 0.72 & 0.39 & 0.00 & 0.24 & 0.75 & 8.55 \\
\hline
\end{tabular}
}
\end{table}

\subsection{Comparing LLM based Sentence Simplification Strategies}

Table~\ref{tab:sentence-simplification-comparison} presents a comparison of sentence-level text simplification performance using two strategies: basic LLM-based simplification and LLM-based plan-driven simplification. Across all metrics, the plan-driven approach yields marginal improvements, particularly in BLEU and FKGL, indicating better fluency and readability while preserving fidelity to the source.

\begin{table}[h!]
\centering
\caption{Comparison of Sentence-Level Simplification Metrics between Basic and Plan-Driven LLM Approaches (Appendix~\ref{app:codabench-task1.1})}
\label{tab:sentence-simplification-comparison}
\begin{tabular}{|l|c|c|c|c|}
\hline
\textbf{Method} & \textbf{SARI\footnotemark[1]} & \textbf{BLEU} & \textbf{BERTScore\_F1} & \textbf{FKGL} \\
\hline
Basic LLM Simplification & 42.887 & 26.4049 & 0.9005 & 9.5452 \\
Plan-Driven LLM Simplification & \textbf{42.985} & \textbf{30.5769} & \textbf{0.9014} & \textbf{9.047} \\
\hline
\end{tabular}
\end{table}

\subsection{Comparing LLM based Document Simplification Strategies}

Table~\ref{tab:summary-guided-vs-direct} presents a comparison between direct LLM-based document simplification and summary-guided simplification across a range of evaluation metrics.

\begin{table}[h!]
\centering
\caption{Performance Comparison: Summary-Guided vs. Direct LLM Document Simplification (Appendix~\ref{app:codabench-task1.2})}
\label{tab:summary-guided-vs-direct}
\begin{tabular}{|l|c|c|c|c|c|}
\hline
\textbf{Method} & \textbf{SARI\footnotemark[1]} & \textbf{BLEU} & \textbf{BERTScore\_F1} & \textbf{FKGL} & \textbf{Token Length} \\
\hline
Direct LLM Simplification & \textbf{43.775} & \textbf{44.6724} & \textbf{0.8605} & \textbf{11.0827} & 257.00 \\
Summary-Guided Simplification & 42.916 & 31.6618 & 0.8493 & 11.2496 & \textbf{249.93} \\
\hline
\end{tabular}
\end{table}
\footnotetext[1]{Only SARI is the official evaluation metric for Task 1. Other metrics are reported for supplementary analysis.}

While the direct LLM-based simplification approach slightly outperforms the summary-guided method in terms of standard metrics such as SARI, BLEU, and BERTScore, the summary-guided approach offers distinct benefits. By first generating a high-level summary and then using it to steer the simplification process, this method can produce more coherent and purpose-driven simplifications. The separation of summarization and simplification stages allows the model to better focus on key ideas, potentially reducing redundancy and irrelevant elaborations.

Moreover, the reduced token length in the summary-guided method suggests more concise output, which can be beneficial in applications where brevity and focus are important. Although the readability score (FKGL) is slightly higher, indicating marginally more complex language, the summary-guided approach may improve factual alignment and structural cohesion.

% \section{Discussion} \label{discussion}

% Discussion of the results and their implications.

% \section{Future Work} \label{future work}

% What would you do next?

\section{Conclusions} \label{conclusions}

This study explores the effectiveness of large language models (LLMs) in text simplification across both sentence and document levels. At the sentence level, our plan-driven approach, which prompts the model to select an explicit simplification strategy before generation, yields improved performance in fluency and readability compared to direct simplification. At the document level, we propose a summary-guided simplification pipeline that, while slightly underperforming in standard metrics, offers qualitative advantages in conciseness and coherence by leveraging intermediate summarization as contextual scaffolding.

Our work demonstrates that incorporating structural planning and summarization can enhance LLM-based simplification, especially for longer and more complex texts. Future research will focus on improving the structural prompting framework by introducing an iterative loop that refines prompts based on automatic evaluation metrics, serving as a built-in feedback mechanism.

\section*{Acknowledgements}

We thank the Data Science at Georgia Tech (DS@GT) CLEF competition group for their support.
This research was supported in part through research cyberinfrastructure resources and services provided by the Partnership for an Advanced Computing Environment (PACE) at the Georgia Institute of Technology, Atlanta, Georgia, USA \cite{PACE}. 

%% The declaration on generative AI comes in effect
%% https://ceur-ws.org/GenAI/Policy.html
\section*{Declaration on Generative AI}

 During the preparation of this work, the authors used ChatGPT and Gemini for grammar and spelling check, as well as assistance in the code for the conducted experiments.  After using these tools, the authors reviewed and edited the content as needed and take full responsibility for the publication’s content. 